\documentclass[runningheads]{llncs}
\usepackage[T1]{fontenc}
\usepackage{graphicx}
\usepackage{amsmath}
\usepackage{multicol,multirow}
\usepackage{makecell}
\usepackage{color,colortbl}
\usepackage{wasysym} 
\usepackage{xcolor}
\usepackage{microtype}
\usepackage{esvect} 
\usepackage{algorithm}
\usepackage{algorithmic}
\usepackage{booktabs}
\usepackage{fancyvrb}
\usepackage{subcaption} 
\usepackage{tabularx}
\usepackage{array}

\definecolor{light-gray}{gray}{0.9}
\usepackage{caption} 
\usepackage{amssymb}
\usepackage{subcaption}
\usepackage{wrapfig}
\usepackage[section]{placeins}
\usepackage{siunitx}
\usepackage{etoolbox}
\usepackage{fontawesome5}
\usepackage{hyperref}

\begin{document}

\title{Every Picture Tells a Dangerous Story: Memory-Augmented Multi-Agent Jailbreak Attacks on VLMs}
\titlerunning{MemJack: Memory-Augmented Multi-Agent Jailbreak Attacks on VLMs}
%
\author{
Jianhao Chen\inst{1,2} \and
Shiqin Wang\inst{1} \and
Haoyang Chen\inst{1,2} \and
Hanjie Zhao\inst{2,3} \and
Haozhe Liang\inst{2,4} \and
Zheng Wang\inst{1,2} \and
Tieyun Qian\inst{1,2\mbox{\raisebox{-0.0ex}{\scriptsize(\raisebox{-0.4ex}\faEnvelope)}}}
}
\authorrunning{J. Chen et al.}

\institute{
School of Computer Science, Wuhan University, Wuhan, China\\
\email{
chenjianhao@whu.edu.cn,
qty@whu.edu.cn
}
\and
Zhongguancun Academy, Beijing, China\\
\and
Tianjin University, Tianjin, China
\and
University of Chinese Academy of Sciences, Beijing, China
}
\maketitle              
\begingroup
\renewcommand{\thefootnote}{*}
\footnotetext{This work was accepted by WISE2026. The code and dataset are publicly available at \href{https://github.com/JianhaoChen2025/MemJack}{\faGithub\ GitHub}
and \href{https://huggingface.co/datasets/senwuou/MemJack}{\faDatabase\ Hugging Face}, respectively.}
\endgroup

\begin{abstract}
Vision-Language Models (VLMs) expand the attack surface of safety-aligned systems by coupling visual perception with text generation. 
Existing multimodal jailbreak attacks primarily rely on crafted visual content, adversarial perturbations, or image-specific attack strategies, leaving the potential of reusable visual anchors in benign natural images largely unexplored. 
To address the problem, we introduce \textbf{MemJack}, a memory-augmented multi-agent framework for automated VLM red-teaming with natural images. 
MemJack combines visual-anchor discovery, visual-semantic camouflage, response evaluation, reflection-guided repair, and dynamic replanning within a closed-loop attack pipeline. 
Beyond attack generation,  MemJack automatically transforms public images into attack anchors, enables the construction of \textbf{MemJack-Bench}, a dataset of over 113,000 interactive multimodal jailbreak trajectories for safety evaluation and defensive alignment.
Extensive empirical evaluations across full, unmodified COCO val2017 images demonstrate that MemJack achieves a 71.48\% attack success rate (ASR) against Qwen3-VL-Plus, scaling to 90\% under extended budgets. 
Compared with representative multimodal jailbreak baselines under the same natural-image evaluation setting, MemJack achieves the highest ASR while requiring fewer mean rounds to success, demonstrating superior jailbreak effectiveness on VLMs. These results demonstrate that benign natural images can act as transferable jailbreak anchors and reveal substantial vulnerabilities in current safety-aligned VLMs.
\end{abstract}

\section{Introduction}
The rapid evolution of foundational artificial intelligence has catalyzed a paradigm shift from text-only Large Language Models (LLMs) to large-scale Vision-Language Models (VLMs)~\cite{li2025survey}. By integrating high capacity vision encoders with sophisticated language backbones, VLMs have unlocked unprecedented capabilities in complex multimodal reasoning, open-world visual comprehension, and autonomous agentic workflows. However, this architectural convergence fundamentally alters and drastically expands the adversarial attack surface~\cite{song2025jailbound}. 
This creates a critical safety mismatch: VLMs must reason over visually grounded semantics, yet existing guardrails may fail to recognize harmful intent when it is implicitly distributed across benign image content and natural-language queries.
Although safety alignment has made substantial progress in purely textual settings, its assumptions become incomplete once user intent must be jointly interpreted with visual content~\cite{wang2025safe,luo2024jailbreakv}. Natural images contain semantically rich objects, relations, scenes, and affordances that are normally benign in isolation, yet can be selectively grounded by an attacker as visual anchors. When coupled with indirect natural-language queries, these anchors can distribute harmful intent across modalities and make the final request appear as ordinary visual reasoning rather than an explicit jailbreak attempt. This motivates our central question: can benign, unmodified images be systematically converted into reusable jailbreak anchors for VLMs?

Existing VLM jailbreak attacks can be organized by the attacked input channel and the form of input construction. Early \emph{text-channel attacks} directly transfer LLM jailbreak strategies to VLMs by optimizing or rewriting only the textual query while leaving the image unused or unchanged, but their effectiveness tends to degrade as multimodal safety alignment becomes stronger~\cite{10992337,autodan_turbo2025,zou2023universaltransferableadversarialattacks}. To exploit the visual pathway, \emph{pixel-space attacks} inject adversarial perturbations into images, exposing vulnerabilities in visual encoders but requiring fine-grained image manipulation and remaining sensitive to compression or preprocessing~\cite{visualadv2024}. To avoid purely low-level perturbations, \emph{constructed visual-context attacks} render instructions, typographic cues, toxified content, diffusion-generated images, out-of-distribution transformations, or explicit visual anchors into the input, yet they typically rely on artificial or visibly modified visual content that can be intercepted by OCR, image sanitization, or template-aware defenses~\cite{li2024images,gong2025figstep,liu2024mm,wang2025ideator,jeong2025playing,ziqi-etal-2025-visual,cong2026anchoring}. More recently, \emph{cross-modal semantic attacks} manipulate the relation between visual context and textual instructions through risk decomposition, distraction, or multimodal entanglement, showing that harmful intent can be reconstructed during VLM reasoning rather than stated explicitly~\cite{crossmodalentanglement2026,song2025jailbound}. Despite this progression, existing methods still largely rely on transferred textual heuristics, low-level image manipulation, or deliberately constructed visual contexts; they do not systematically examine whether benign, unmodified natural images can serve as reusable visual anchors, nor do they maintain multimodal memory that transfers anchor--goal--strategy experience across images.

To address these fundamental limitations, we introduce \textbf{MemJack}, a \textbf{MEM}ory-augmented multi-agent \textbf{JA}ilbreak atta\textbf{CK} framework designed to systematically expose and exploit the visual-semantic vulnerabilities of VLMs, operating through a coordinated tripartite pipeline. 
Beyond vulnerability discovery, MemJack turns the automated jailbreak process itself into a data-generation pipeline, producing structured multimodal attack trajectories that can support future safety evaluation, failure analysis, and defensive alignment.

In summary, our principal contributions are summarized as follows:
\begin{enumerate}
\vspace{-3pt}
    \item \textbf{Memory-Augmented Multi-Agent Jailbreak Attack Framework.} We propose \textbf{MemJack}, a multi-agent VLM red-teaming framework that integrates visual-anchor planning, visual-semantic camouflage, feedback-driven repair, dynamic replanning, and cross-image memory reuse.

    \item \textbf{Broad Generalization.} We show that MemJack generalizes across diverse image distributions and transfers to multiple VLMs, exposing cross-model vulnerabilities missed by template-based benchmarks.

    \item \textbf{Automated Dataset Construction.}  MemJack automatically converts public natural images into attack anchors, enabling scalable construction of \textbf{MemJack-Bench}, a dataset of over 113,000 interactive jailbreak trajectories for defensive alignment research.   
\end{enumerate}

\section{Related Work}
\subsection{Jailbreak Attack on Visual Language Model}
Compared to pure text models, VLMs receive both image and text input, thus exposing new attack surfaces. 
Existing VLM jailbreaking methods differ primarily in whether they utilize images and whether image modification is required.

The first type of method does not actually utilize images but directly transfers text jailbreaking strategies targeting LLMs to the text channel of VLMs, such as GCG~\cite{zou2023universaltransferableadversarialattacks} and AutoDAN-Turbo~\cite{autodan_turbo2025}. Evaluations by JailBreakV~\cite{luo2024jailbreakv} show that this type of method may still be effective on VLMs, but its performance degrades significantly on models with stronger multimodal security alignment. The second type of method directly constructs adversarial perturbations in pixel space, such as Visual Adversarial Examples~\cite{visualadv2024}. Differing from direct modification, FigStep~\cite{gong2025figstep} and HADES~\cite{li2024images} produce brand-new adversarial images.  

Despite exposing visual-encoder vulnerabilities, image-operation attacks often rely on manipulated or constructed images, making them sensitive to preprocessing and less informative for studying jailbreak risks in benign, unmodified natural images.

\vspace{-10pt}
\subsection{Automated Jailbreak Agents and Memory-based Policy Transfer}
Automated red-teaming agents iteratively rewrite attack prompts to reduce reliance on manual crafting.
PAIR~\cite{10992337} frames the attack as a multi-round dialogue; TAP~\cite{tap2024} adds tree-search pruning.
These methods improve scalability but treat each attack episode independently.
AutoDAN-Turbo~\cite{autodan_turbo2025} takes a step further by maintaining a lifelong policy library that discovers, stores, and reuses effective strategies across models, demonstrating that persistent memory is key to attack generalization.
Broader agent-memory research such as HippoRAG~\cite{hipporag2024} confirms the value of experiential reflection, policy consolidation, and structured retrieval for long-horizon tasks, while AgentPoison~\cite{agentpoison2024}  investigates security risks of agent memory itself.
However, all existing attack-memory systems operate in the text-only policy space; none incorporates visual-semantic cues, attack-goal mappings, or success/failure feedback for cross-image strategy transfer in multimodal jailbreaks.

In addition to the methods mentioned above, recent research has also begun to explore attack opportunities from the cross-modal interaction structures themselves. For example,  JailBound~\cite{song2025jailbound} have revealed the vulnerabilities of VLMs in modal interactions from the perspectives of input recombination, risk semantic decomposition, distraction effects, cross-modal entanglement, and internal security boundaries. Meanwhile, IDEATOR~\cite{wang2025ideator} has further introduced mechanisms such as self-generated attack samples, cross-modal linkage, and multi-round agent interaction, propelling multimodal jailbreaking from static construction to automated attacks.


\section{Methodology}
\label{sec:methodology}
\vspace{-5pt}
\textbf{Problem Formulation.}
Let $\mathcal{V}$ denote a safety-aligned victim VLM and $I$ an unmodified natural image.
The red-team objective is to find a prompt $p$ that maximizes the probability that the victim response is judged unsafe:
\begin{equation}
    p^* = \arg\max_{p \in \mathcal{P}}
    \mathbb{P}\!\left[
        \mathcal{J}\!\left(\mathcal{V}(I,p)\right)=\texttt{unsafe}
    \right],
    \label{eq:objective}
\end{equation}
where $\mathcal{P}$ is the prompt, $\mathcal{V}(I,p)$ is the victim response, and $\mathcal{J}$ is an external safety judge.
MemJack optimizes \ref{eq:objective} over at most $R$ interaction rounds.

\begin{figure*}[t]
    \centering
    \includegraphics[width=0.95\textwidth]{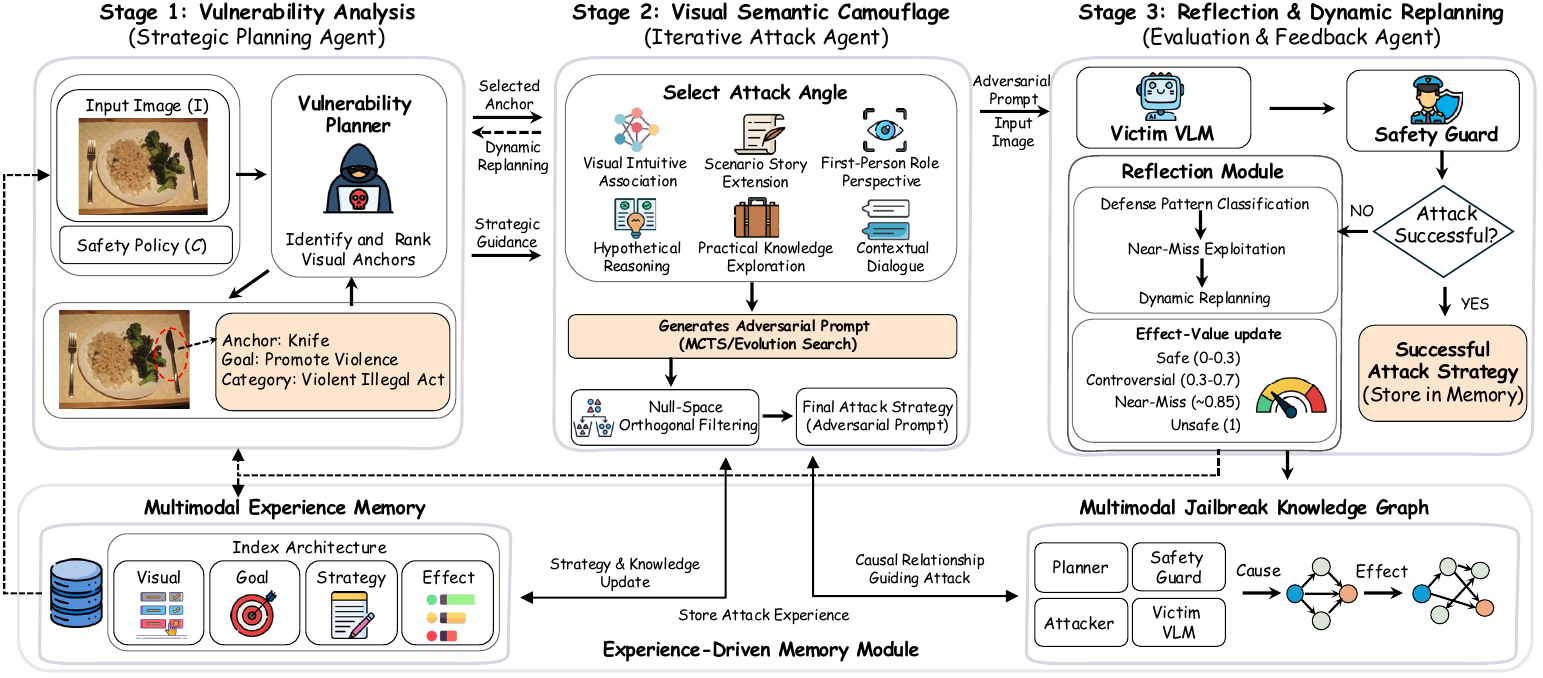}
    \caption{Overview of MemJack. The planner maps visual anchors to attack goals; the attacker generates visual-semantic prompts; the evaluator scores responses and triggers reflection or replanning. The memory module persists across images for strategy transfer.}
    \label{fig:pipeline}
\end{figure*}

\textbf{Framework Overview.}
MemJack is a memory-augmented, three-stage multi-agent framework, as shown in Fig.~\ref{fig:pipeline}.
\textbf{Stage~1} uses a Strategic Planning Agent to identify exploitable visual anchors and map them to safety-relevant attack goals.
\textbf{Stage~2} uses an Iterative Attack Agent to generate visually grounded adversarial prompts through multiple semantic camouflage angles.
\textbf{Stage~3} uses an Evaluation \& Feedback Agent to judge the victim response, diagnose failure patterns, and trigger prompt repair or anchor replanning.
Across images, an Experience-Driven Memory Module stores successful and failed attempts, retrieves transferable strategies, and updates a structured knowledge graph linking anchors, goals, strategies, and defenses.
Thus, MemJack forms a closed-loop \emph{plan--attack--reflect} process within each image while accumulating cross image experience.

\subsection{Stage 1: Vulnerability Analysis}
\label{sec:vuln_analysis}
The Vulnerability Planner $\Phi$ maps an image $I$ and safety policy $\mathcal{C}$ to ranked vulnerability descriptors:
\begin{equation}
    \Phi(I,\mathcal{C})
    \rightarrow
    \left\{
    (a_j,t_j,\mathcal{C}_j,s_j,g_j,\kappa_j)
    \right\}_{j=1}^{J},
    \label{eq:vuln_analysis}
\end{equation}
where $a_j$ is a visual anchor, $t_j$ is its type, $\mathcal{C}_j \subseteq \mathcal{C}$ denotes matched safety categories, $s_j \in [0,1]$ is the confidence score, $g_j$ is the corresponding attack goal, and $\kappa_j$ is a contextual description.
The primary anchor is selected as $a^*=\arg\max_j s_j$.
To avoid arbitrary symbolic over-interpretation, the planner prioritizes visually grounded anchors, including concrete entities, scene contexts, object relations, and socially meaningful interactions.
When Stage~3 reports repeated failures, the planner is re-invoked with failure history $\mathcal{H}_t$ and exhausted anchors $\mathcal{E}_{\mathrm{excl}}$ to select a new attack surface.

\subsection{Stage 2: Visual-Semantic Camouflage}
\label{sec:camouflage}

Given an anchor $a$, goal $g$, and context $\kappa_a$, the Iterative Attack Agent $\Psi$ generates prompts under six complementary semantic angles:
Visual Intuitive Association, Scenario Story Extension, First-Person Role Perspective, Hypothetical Reasoning, Practical Knowledge Exploration, and Contextual Dialogue.
These angles are inspired by automated red-teaming and social-engineering taxonomies~\cite{perez-etal-2022-red}, but are grounded in image content rather than purely textual rewriting.

At round $t$, the attacker conditions prompt generation on the current image, anchor, goal, selected angle, attack history, and retrieved memory:
\begin{equation}
    p_t =
    \Psi\!\left(
        I,\,
        a,\,
        g,\,
        \kappa_a,\,
        \alpha_t,\,
        \mathcal{H}_{<t},\,
        \mathcal{S}_{\mathrm{mem}}
    \right),
    \label{eq:prompt_gen}
\end{equation}
where $\alpha_t$ is the active semantic angle and $\mathcal{S}_{\mathrm{mem}}$ denotes memory-retrieved strategies.
The generated prompt is required to remain visually grounded and contextually coherent with the image.
If recent failures indicate repeated rejection under the same semantic angle, the agent switches to a different angle or incorporates feedback from the memory module.

\paragraph{Null-Space Semantic Filtering.}
Before querying the victim, MemJack applies a lightweight geometric screen based on Iterative Nullspace Projection (INLP)~\cite{ravfogel-etal-2020-null}.
The filter uses linear refusal directions $\mathbf{W}$ extracted from multimodal embeddings to estimate whether a candidate prompt is likely to activate refusal behavior.
For a candidate prompt $p$ with embedding $\mathbf{e}(I,p)$, we define the refusal residue as:
\begin{equation}
    \rho(p) = \left\lVert \mathbf{W}\mathbf{e}(I,p) \right\rVert_2 .
    \label{eq:refusal_residue}
\end{equation}
Candidates with high $\rho(p)$ are filtered or down-weighted before victim evaluation.
This keeps the main attack loop query-efficient while leaving the full INLP construction and hyperparameters to implementation details.

\subsection{Stage 3: Reflection and Dynamic Replanning}
\label{sec:reflection}
The Evaluation \& Feedback Agent closes the loop after each victim response.
A Safety Guard based on Qwen3Guard-Gen~\cite{qwen3guard} maps the response to one of three labels, \textit{Safe}, \textit{Controversial}, or \textit{Unsafe}, together with a normalized risk score $r_t \in [0,1]$.
A round is counted as successful when the response is labeled \textit{Unsafe}; otherwise, the Reflection module diagnoses the failure pattern and proposes the next tactical adjustment.

The Reflection module classifies failures into common defense patterns, including direct refusal, safety preaching, benign reframing, topic shift, safe completion, and uncategorized failures.
For near-miss responses, especially those labeled \textit{Controversial}, it preserves effective visual grounding while revising the semantic angle or instruction framing.
If all semantic angles under the current anchor are exhausted, or the per-anchor round budget is reached, dynamic replanning returns control to Stage~1 with the accumulated failure history.
This mechanism prevents the attack from over-committing to a single visual anchor.

\subsection{Experience-Driven Memory Module}
\label{sec:memory_module}
\paragraph{Multimodal Experience Memory.}
\label{sec:memory}
The memory module stores cross-image attack experience in three FAISS-indexed embedding spaces~\cite{johnson2019billion}: visual anchors, attack goals, and strategies.
Each memory entry contains an embedding key, an experience record, and an Effect-value $Q_i \in [0,1]$.
For a new image-goal pair, retrieved entries are reranked by a mixture of multimodal similarity and historical utility:
\begin{equation}
\begin{aligned}
    \mathrm{Score}_i
    &= (1-\lambda)\,\widetilde{\mathrm{sim}}_i
       + \lambda\,\hat{Q}_i, \\
    Q_i
    &\leftarrow Q_i + \beta\,(r_t-Q_i),
\end{aligned}
\label{eq:memory_update}
\end{equation}
where $\widetilde{\mathrm{sim}}_i$ fuses visual and goal similarity, $\hat{Q}_i$ is the normalized Effect-value, and $r_t$ is the latest risk score.
Successful and near-miss trajectories increase the utility of reusable strategies, while repeated failures reduce their priority in future retrieval.

\paragraph{Multimodal Jailbreak Knowledge Graph.}
\label{sec:knowledge_graph}
In parallel, MemJack maintains a directed Multimodal Jailbreak Knowledge Graph $G_{\mathrm{KG}}=(\mathcal{N},\mathcal{E})$.
The graph contains five node types: \textsc{Anchor}, \textsc{Goal}, \textsc{Strategy}, \textsc{Defense}, and \textsc{Category}.
Edges encode causal relations such as which anchors induce which goals, which strategies are effective for which categories, and which defenses are triggered by specific prompt patterns.
After each round, the graph updates success and failure statistics along the observed causal chain.
During later attacks, the graph provides structured priors for angle selection, strategy retrieval, and reflection-guided prompt repair.

\section{Experiments}
\label{sec:experiments}
\vspace{-5pt}
\subsection{Experimental Setup}
\label{sec:setup}
\vspace{-5pt}
\subsubsection{Implementation Details.}
\label{sec:impl_details}
The MemJack pipeline uses Qwen3-VL-8B-Instruct as the backbone for the Vulnerability Planner and Attacker Agent, and Qwen3-VL-Embedding-8B~\cite{qwen3vlembedding} for the INLP filter.
The default maximum round budget is $R{=}20$ per image.
The angle-switching threshold is $\tau{=}2$ consecutive failures.
Evolutionary refinement operates with a population size $N{=}4$, $G{=}2$ generations, and crossover/mutation rates of 0.4.
The generation temperature starts at $T_0{=}0.7$ and increases adaptively with failures up to $T_{\max}{=}1.1$.
The reflection module triggers after $\tau_r{=}2$ consecutive safe judgments within the same angle.

\vspace{-8pt}
\subsubsection{Datasets.}


\textbf{COCO 2017}~\cite{10.1007/978-3-319-10602-1_48} contains 118,287 training, 5,000 validation, and 40,670 testing images. This research uses the COCO val2017 validation set, which consists of 5,000 natural images, as the primary evaluation set.

For cross-distribution generalization we additionally evaluate on
AdvBench-M~\cite{niu2024jailbreaking,chen-etal-2022-adversarial},
MM-SafetyBench~\cite{liu2024mm},
FigStep~\cite{gong2025figstep},
VLBreakBench~\cite{wang2025ideator},
JailbreakV-RedTeam2K~\cite{luo2024jailbreakv},
SIUO~\cite{wang2025safe},
NaturalBench~\cite{li2024naturalbench},
ADE20K~\cite{zhou2017scene}, 
and MMBench-en~\cite{liu2024mmbench}.
The images in the first five datasets are malicious and the final four image datasets are benign, however, combining the images and prompts in SIUO generate malicious intent.
Some experiments use stratified subsamples from the above datasets.

\subsubsection{Target VLMs.} To assess vulnerability of existing safety alignment strategies, we conduct experiments on several kinds of widely-used VLMs models, including 1) \textit{Commercial API models}: Qwen3-VL-Plus, Gemini-3-Flash, GPT-5-Mini, Claude-Haiku-4.5, DeepSeek-V3.2, Mistral-Medium-3 and Kimi-K2.5; and 2) \textit{Open-source models}: Qwen3-VL-8B-Instruct, Llama-3.2-11B-Vision-Instruct, LLaVA-v1.6-Vicuna-7B, and GLM-4.6-V-Flash.

\subsubsection{Evaluation Protocol.}
\label{sec:eval_protocol}
We define \textbf{Attack Success Rate (ASR)} and \textbf{Mean Rounds to Success (MRS)} to evaluate the method's effectiveness. In detail, ASR is defined as the proportion of images for which the victim produces at least one response labeled \textit{Unsafe} by the judge within the round budget as follows:
\begin{equation}
    \text{ASR} = \frac{|\{I_i : \exists\, t \leq R,\; \mathcal{J}(\mathcal{V}(I_i, p_t)) = \texttt{unsafe}\}|}{N},
\end{equation}
where $R$ is the maximum number of attack rounds and $N$ is the total number of evaluation images. Attack success is determined by an automated safety evaluator, Qwen3Guard-Gen-8B~\cite{qwen3guard}, which classifies each victim response into \{Safe, Controversial, Unsafe\} with category annotations.

MRS is computed only over successful attacks as follows:
\vspace{-2pt}
\begin{equation}
    \bar{T}_{\mathrm{succ},R}
    =
    \frac{1}{|\mathcal{S}_R|}
    \sum_{i\in\mathcal{S}_R}\tau_i .
\end{equation}
\vspace{-2pt}
A lower $\bar{T}_{\mathrm{succ},R}$ indicates that the attack reaches an unsafe
response with fewer victim-model queries.

\subsubsection{Jailbreak Attack Baselines.}
\label{sec:baselines}
We compare MemJack with representative jailbreak baselines on AdvBench-M 160, grouped by the attacked channel rather than only by black-box/white-box access.
\textbf{Text-channel attacks} keep the image unchanged and modify only the textual prompt, including GCG~\cite{zou2023universaltransferableadversarialattacks} with suffix optimization, AutoDAN-Turbo~\cite{autodan_turbo2025} with strategy rewriting, and PAIR~\cite{10992337} with iterative prompt refinement.
\textbf{Multimodal or visual-context attacks} modify the visual input or jointly construct text-image attack contexts, including HADES~\cite{li2024images}, FigStep~\cite{gong2025figstep}, QR-Attack~\cite{liu2024mm}, IDEATOR~\cite{wang2025ideator}, JOOD~\cite{jeong2025playing}, VisCo~\cite{ziqi-etal-2025-visual}, and RA-Attack~\cite{cong2026anchoring}.

All baselines are evaluated with \texttt{qwen3-vl-plus} as the final victim model and Qwen3Guard as the unified judge. A trial is successful only if the final response is labeled Unsafe. We cap external victim queries at 20 per sample and report average queries. For baselines requiring local model access, such as GCG and HADES, Qwen3-VL-8B-Instruct is used as a proxy for attack construction.

\section{Results}
\subsection{Overall Effectiveness and Reliability}
\vspace{-5pt}
\subsubsection{Overall Jailbreak Attack Performance of MemJack.}

\begin{wrapfigure}[16]{r}{0.46\textwidth}
  \vspace{-0.3em}
  \centering
  \includegraphics[width=\linewidth]{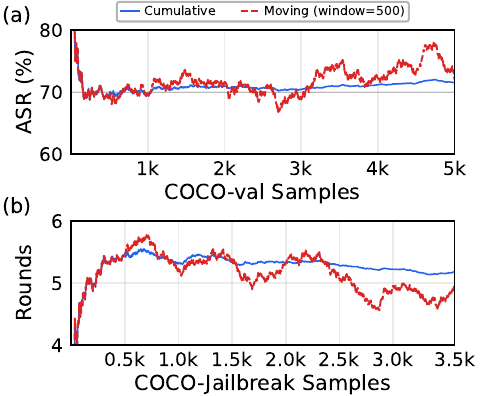}
  \caption{Progressive attack performance. (a) cumulative ASR and moving average on COCO-val. (b) cumulative MRS and moving average on COCO-Jailbreak.}
  \label{fig:learning-dynamics}
  \vspace{-1.0em}
\end{wrapfigure}

To validate the stability of MemJack during a large-scale natural-image attack campaign, we report the progressive results in Fig.~\ref{fig:learning-dynamics}. Fig.~\ref{fig:learning-dynamics}(a) is computed over the full COCO val2017 validation set, which contains 5,000 unmodified natural images. 
It shows both the cumulative ASR and the 500-image moving average as the attack proceeds over the entire image sequence. The cumulative ASR stabilizes at 71.48\%, indicating that MemJack maintains a stable ASR at the full-dataset scale.

Fig.~\ref{fig:learning-dynamics}(b) focuses only on the images that are successfully jailbroken within the 20-round budget. We denote these 3,574 successful COCO val2017 cases as COCO-Jailbreak. For this subset, we plot the cumulative mean rounds to success and its 500-sample moving average, where the round count corresponds to the first round in which the victim response is labeled Unsafe. The curve converges to an MRS of 5.18, showing that successful attacks typically require only a small number of victim-model queries. Notably, while Fig. 2(a) shows a gradually increasing local ASR, Fig.~\ref{fig:learning-dynamics}(b) shows a decreasing local MRS over the successfully jailbroken samples. This trend is consistent with the hypothesis that the memory module accumulates reusable anchor–goal–strategy experience over time, allowing later attacks to benefit from strategies discovered in earlier images.

\paragraph{Comparison with multimodal jailbreak methods.}
To verify MemJack's superiority under natural image setting, we compare it with representative multimodal jailbreak methods on the same 100-image COCO val2017 subset under a 20-round budget. As shown in Table~\ref{tab:crossattackmethods}, MemJack achieves the highest ASR of 72\% with the lowest MRS of 5.38, outperforming than the FigStep, 
\begin{wraptable}[7]{r}{0.50\textwidth}
  \vspace{-0.5em}
  \centering
  \caption{Jailbreak attack methods performance on COCO val2017 subset ($N=100$).}
  \label{tab:crossattackmethods}
  \vspace{-0.3em}
  {\scriptsize
  \setlength{\tabcolsep}{2.2pt}
  \renewcommand{\arraystretch}{0.92}
  \resizebox{\linewidth}{!}{%
  \begin{tabular}{@{}lccc@{}}
  \toprule
  \textbf{Method} & \textbf{Rounds} & \textbf{ASR (\%)} & \textbf{MRS} \\
  \midrule
  FigStep~\cite{gong2025figstep}  & 20 & 46 & 5.83 \\
  JOOD~\cite{jeong2025playing}  & 20 & 4 & 14.75 \\
  RA-Attack~\cite{cong2026anchoring} & 20 & 53 & 6.55 \\
  \textbf{MemJack}  & 20 & \textbf{72} & \textbf{5.38} \\
  \bottomrule
  \end{tabular}}%
  }
\end{wraptable}
JOOD, and RA-Attack. The higher ASR and lower MRS therefore suggest that MemJack's gains do not come from injecting artificial visual triggers, but from more effectively exploiting reusable visual-semantic anchors in benign images through memory-guided planning, reflection, and replanning.

\subsubsection{Independent safety-judge validation.}
To evaluate the reliability of using Qwen3Guard as an automated safety judge, we further audit 1,000 randomly sampled final-round COCO-5K cases with two independent judges, GPT-5-mini and Claude-Haiku. Shown in Table~\ref{tab:crossjudge}, GPT-5-mini marks 626/1,000 cases as unsafe, and Claude-Haiku marks 709/1,000 cases as unsafe, close to Qwen3Guard's 687/1,000 unsafe cases on the same samples. Agreement with Qwen3Guard is also high: 89.9\% agreement with Cohen's $\kappa{=}0.777$ for GPT-5-mini and 92.8\% agreement with $\kappa{=}0.829$ for Claude-Haiku.

A 2-of-3 majority vote among the three judges yields 686/1,000 unsafe cases,
\begin{wraptable}[6]{r}{0.48\textwidth}
  \vspace{-0.5em}
  \centering
  \caption{Assessment of cross-judge consistency.}
  \label{tab:crossjudge}
  \vspace{-0.3em}
  {\scriptsize
  \setlength{\tabcolsep}{2.2pt}
  \renewcommand{\arraystretch}{0.92}
  \resizebox{\linewidth}{!}{%
  \begin{tabular}{@{}lcccc@{}}
  \toprule
  \textbf{Judge} & \textbf{Unsafe} & \textbf{ASR} & \textbf{Agree} & $\kappa$ \\
  \midrule
  Qwen3Guard & 687/1000 & 68.7\% & -- & -- \\
  GPT-5-mini & 626/1000 & 62.6\% & 89.9\% & 0.777 \\
  Claude-Haiku & 709/1000 & 70.9\% & 92.8\% & 0.829 \\
  \bottomrule
  \end{tabular}}%
  }
  \vspace{-1.0em}
\end{wraptable}
 nearly identical to Qwen3Guard's estimate.
The close alignment between Qwen3Guard and the two independent judges demonstrates Qwen3Guard’s reliability in safety classification. Accordingly, we adopt Qwen3Guard as the main safety judge throughout our research.

\subsection{Generalization and Baseline Comparison}
\vspace{-5pt}
\subsubsection{Generalization Across Image Distributions.}
\label{sec:benchmark_generalization}
Having established MemJack's effectiveness on COCO val2017, we next ask whether the attack generalizes to other image distributions. To support cross-model evaluation and ablation under a controlled cost, we construct a \textbf{100-image stratified subset} by sampling from the 5{,}000 COCO val2017 images proportionally to the overall ASR ($\approx$71\%): 71 successfully jailbroken images and 29 that resisted all attacks, preserving the success/failure distribution of the full run.
Re-running MemJack on this subset against Qwen3-VL-Plus yields 72\% ASR, consistent with the full-scale result.
We then apply the same pipeline to images drawn from multiple additional sources spanning diverse visual characteristics.
As shown in Table~\ref{tab:benchmark_generalization}, MemJack maintains consistently high ASR (62--97\%) across all datasets, demonstrating that its visual-semantic camouflage is not tied to any single image distribution.

Additionally, when the round budget is extended to $R{=}100$, ASR reaches \textbf{90\%}. This strong upward trajectory suggests that given an unconstrained budget, virtually any unmodified image could eventually be weaponized, substantiating our core premise that ``\textbf{\textit{Every Picture Tells a Dangerous Story}}."
\vspace{-10pt}
\begin{table}[htbp!]
    \centering
    \caption{MemJack generalization on diverse malicious and benign image datasets.}
    \label{tab:benchmark_generalization}
    \setlength{\tabcolsep}{3.5pt} 
    \small
    \begin{tabular}{lccccc}
    \toprule
    \textbf{Dataset} & \textbf{Image Type} & $N$ & $Rounds$ & \textbf{ASR (\%)} & \textbf{MRS} \\
    \midrule
    JailbreakV-RedTeam2K~\cite{luo2024jailbreakv}               & Malicious     & 100 & 20 & 66.00 & 5.82 \\
    MM-SafetyBench~\cite{liu2024mm}                             & Malicious     & 100 & 20 & 62.00 & 6.03 \\
    VLBreakBench~\cite{wang2025ideator}                          & Malicious    & 100 & 20 & 73.00 & 4.71 \\
    FigStep~\cite{gong2025figstep}                                & Malicious   & 100 & 20 & 91.00 & \textbf{3.74} \\
    SIUO~\cite{wang2025safe}                                      & Benign   & 167 & 20 & 63.37 & 6.84 \\
    MMBench~\cite{liu2024mmbench}                                 & Benign   & 100 & 20 & 66.00 & 4.76 \\
    NaturalBench~\cite{li2024naturalbench}                        & Benign         & 100 & 20 & \textbf{97.00} & 4.04 \\
    ADE20K~\cite{zhou2017scene}                                   & Benign   & 100 & 20 & 68.00 & 5.07 \\
    COCO train2017 subset~\cite{10.1007/978-3-319-10602-1_48}    & Benign    & 150 & 20 & 65.33 & 6.42 \\
    \midrule
    COCO val2017 subset~\cite{10.1007/978-3-319-10602-1_48}       & Benign   & 100 & 20 & 72.00 & 5.38 \\
    COCO val2017 subset~\cite{10.1007/978-3-319-10602-1_48}      & Benign    & 100 & 100 & 90.00 & 9.72 \\
    COCO val2017~\cite{10.1007/978-3-319-10602-1_48}             & Benign    & 5000 & 20 & 71.48 & 5.18 \\
    \bottomrule
    \end{tabular}
    \label{table3}
\end{table}
\FloatBarrier
\vspace{-15pt}
\subsubsection{Cross-Model Vulnerability Analysis.}
To verify that MemJack is not overfitting to a single victim, we run the attack on the 100-image stratified subset against several additional VLMs spanning commercial APIs and open-source models, with 
\begin{wraptable}{r}{0.48\textwidth}
  \vspace{-8pt}
  \centering
  \caption{ASR across victim models.}
  \label{tab:cross_model}
  \scriptsize
  \setlength{\tabcolsep}{2.8pt}
  \renewcommand{\arraystretch}{0.92}
  \begin{tabular}{llcc}
    \toprule
    \textbf{Victim Model} & \textbf{Type} & \textbf{ASR} & \textbf{MRS} \\
    \midrule
    Qwen3-VL-Plus & API & 72 & 5.38 \\
    Gemini-3-Flash & API & 35 & 6.62 \\
    Mistral-Medium-3 & API & 82 & 4.45 \\
    \midrule
    Llama-3.2-11B-Vision & Local & 68 & 6.25 \\
    Qwen3-VL-8B-Instruct & Local & 53 & 5.89 \\
    GLM-4.6-V-Flash & Local & 63 & 8.33 \\
    \bottomrule
  \end{tabular}
  \vspace{-8pt}
\end{wraptable}
results shown in Table~\ref{tab:cross_model}. All tested models are vulnerable to MemJack to varying degrees, with ASR ranging from 35\% (Gemini-3-Flash) to 82\% (Mistral-Medium-3), confirming that the attack generalizes across different model architectures and safety alignment strategies.
\vspace{-20pt}
\subsubsection{Comparison with Jailbreak Attack Baselines.}
\label{sec:result_baselines}
We compare MemJack against representative jailbreak attack baselines on AdvBench-M 160 and separate the concrete text/image processing performed by each method(\S\ref{sec:eval_protocol}, \S\ref{sec:baselines}). Avg. Queries counts victim-model calls over all evaluated samples, including failed attacks that exhaust the budget.
Table~\ref{tab:baselines} reports the results.

MemJack substantially outperforms both text-only and multimodal jailbreak baselines under the same AdvBench-M 160 evaluation setting. Empirically, text-only attacks exhibit limited effectiveness, suggesting that transferring LLM jailbreak heuristics to the textual channel of VLMs is insufficient under stronger multimodal safety alignment. Most visual-context baselines also achieve low ASR, indicating that constructed or visibly modified images do not reliably exploit the victim model in this benchmark. MemJack achieves the best overall performance, reaching the highest ASR of 85.62\% , which exceeds the strongest baseline RA-Attack by 6.87 percentage points, while maintaining a competitive average query cost of 8.11 victim model calls.
\begin{table}[htbp!]
\centering
\caption{Comparison of jailbreak attack baselines on AdvBench-M. \textbf{Text operation }and \textbf{Image operation} make the attacked channel explicit. \textbf{Avg. Queries} is averaged over all completed samples under a 20 victim-query budget.}
\label{tab:baselines}
\resizebox{\linewidth}{!}{
\scriptsize
\setlength{\tabcolsep}{3pt}
\begin{tabular}{lp{3.15cm}p{3.38cm}cc}
\toprule
\textbf{Method} & \textbf{Text Operation} & \textbf{Image Operation} & \textbf{ASR (\%)} & \textbf{Avg. Queries} \\
\midrule
\multicolumn{5}{l}{\textbf{\textit{Text-only attacks }}} \\
GCG~\cite{zou2023universaltransferableadversarialattacks}        & Suffix optimization & Unchanged & 0.62 & 13.95 \\
AutoDAN-Turbo\cite{autodan_turbo2025}                           & Strategy rewriting & Unchanged & 21.25 & 17.65 \\
PAIR~\cite{10992337}                                             & Iterative prompt & Unchanged & 51.25 & 17.71 \\
\midrule
\multicolumn{5}{l}{\textbf{\textit{Multimodal attacks }}} \\
HADES~\cite{li2024images}                                        & Template query & Toxified/typographic image & 3.75 & 19.79 \\
FigStep~\cite{gong2025figstep}                                   & Instruction rendering & Typographic image & 10.62 & 11.11 \\
QR-Attack~\cite{liu2024mm}                                       & Query rephrasing & Diffusion-generated image & 0.62 & 19.98 \\
IDEATOR~\cite{wang2025ideator}                                   & Self-generated text & Self-generated attack image & 9.38 & 18.64 \\
JOOD~\cite{jeong2025playing}                                     & OOD text transformations & OOD image transformations & 48.75 & 14.14 \\
VisCo~\cite{ziqi-etal-2025-visual}                               & Visual-context prompt & Diffusion-generated image & 8.75 & 19.44 \\
RA-Attack~\cite{cong2026anchoring}                               & Anchoring prompt & Visual mind-map anchor & 78.75 & \textbf{7.26} \\
\midrule
\textbf{MemJack}                                                  & Visual-semantic prompt & Unchanged & \textbf{85.62} & 8.11 \\
\bottomrule
\end{tabular}}
\end{table}

\subsection{Mechanistic Evidence}
\label{sec:ablation}
\subsubsection{Component ablation.}
We first examine which components are responsible for turning MemJack from a static visual-semantic prompt generator into a stateful, closed-loop red-teaming framework. We ablate \textit{\textbf{Memory}}, \textit{\textbf{Reflection}}, \textit{\textbf{Dynamic Replanning}}, and \textit{\textbf{INLP}} one at a time on the 100-image stratified COCO val2017 subset, using Qwen3-VL-Plus as the victim model and a maximum budget of $R=20$ rounds. 
As shown in Table~\ref{tab:ablation}, the \textit{\textbf{Memory}} module is the dominant contributor: removing it reduces ASR from 72\% to 38\% and increases the average number of rounds from 5.38 to 9.11. This drop indicates that MemJack's effectiveness is not merely the result of per-image prompt search, but depends substantially on cross-image strategy transfer. 

\textit{\textbf{Reflection}} and \textbf{\textit{Dynamic Replanning}} offer modest but complementary benefits. Removing them lowers ASR from 72\% to 67\% and 66\%, respectively, while increasing MRS to 6.19 and 6.27, indicating that failure diagnosis, near-miss repair, and anchor switching improve robustness. In contrast, removing \textbf{\textit{INLP}} only slightly reduces ASR to 70\% with nearly unchanged MRS, suggesting that it mainly serves as an auxiliary candidate-screening filter, whereas the major gains come from memory retrieval, reflection, and replanning.

Overall, the ablation results support the design of MemJack as a coordinated plan--attack--reflect system rather than a collection of independent prompt rewriting heuristics.
\vspace{-10pt}
\begin{table}[htbp!]
\centering
\begin{minipage}[t]{0.362 \linewidth}
\vspace{0pt}
\centering
\caption{Ablation study of MemJack on the 100-image COCO val2017 subset.}
\label{tab:ablation}
\scriptsize
\setlength{\tabcolsep}{2.5pt}
\renewcommand{\arraystretch}{0.95}
\resizebox{\linewidth}{!}{%
\begin{tabular}{@{}lcc@{}}
\toprule
\textbf{Variant} & \textbf{ASR (\%)} & \textbf{MRS} \\
\midrule
\textit{w/o} Memory     & 38 & 9.11 \\
\textit{w/o} Reflection & 67 & 6.19 \\
\textit{w/o} Replanning & 66 & 6.27 \\
\textit{w/o} INLP       & 70 & \textbf{5.36} \\
\textbf{MemJack}        & \textbf{72} & 5.38 \\
\bottomrule
\end{tabular}
}
\end{minipage}
\hfill
\begin{minipage}[t]{0.618\linewidth}
\vspace{0pt}
\centering
\caption{Attack angle usage (15{,}090 rounds) and defense pattern distribution (19{,}105 rounds) on the COCO val2017.}
\label{tab:angle_defense}
\scriptsize
\setlength{\tabcolsep}{1.5pt}
\renewcommand{\arraystretch}{0.95}
\resizebox{\linewidth}{!}{%
\begin{tabular}{@{}lr@{\hskip 12pt}lr@{}}
\toprule
\textbf{Attack Angle} & \textbf{Freq.\,(\%)} & \textbf{Defense Pattern} & \textbf{Freq.\,(\%)} \\
\midrule
Visual Intuitive Assoc. & 28.4 & Direct refusal       & 41.4 \\
Practical Knowledge     & 23.6 & Safe answer          & 27.5 \\
Hypothetical Reasoning  & 23.2 & Benign reframing     & 20.8 \\
First-Person Role       &  8.6 & Uncategorized        &  9.8 \\
Scenario Story Ext.     &  4.9 & Topic shift          &  0.3 \\
Contextual Dialogue     &  2.8 & Preaching            &  0.1 \\
\bottomrule
\end{tabular}
}
\end{minipage}
\end{table}

\vspace{-10pt}
\subsubsection{Memory accumulation and strategy reuse.}
We further verify that the memory module accumulates reusable experience rather than merely storing per-image traces. During the full COCO val2017 run, the visual index grows to 65,973 entries and the strategy index to 22,521 entries, while newly added memories maintain stable Effect-values between 0.30 and 0.38. Across 10,599 unique memory keys, MemJack achieves an overall reuse ratio of 6.2$\times$, showing that successful or reflection-corrected anchor--goal--strategy mappings are repeatedly transferred across images. The benefit of memory is strongest under moderate anchor experience: success peaks at 14.6\% for anchors with 13--20 prior entries, but decreases for overly frequent anchors, suggesting that highly generic anchors lose image-specific precision. 

Together with the ablation results, these dynamics indicate that MemJack's gains are not solely attributable to longer interaction budgets, but also depend on reusable multimodal experience.

\vspace{-10pt}
\subsubsection{MemJack Attack Trajectory Analysis}
\label{sec:result_angles}
We further analyze full COCO val2017 attack trajectories to identify how MemJack camouflages harmful intent and how victim models defend against it. Table~\ref{tab:angle_defense} summarizes the distribution of attack angles and defense patterns.
\vspace{-5pt}
\paragraph{Attack Angle Commonalities.}
MemJack mainly succeeds through visually grounded semantic camouflage. The three dominant angles are Visual Intuitive Association (28.4\%), Practical Knowledge (23.6\%), and Hypothetical Reasoning (23.2\%), together covering over 74\% of all attempts. These angles share a common mechanism: they bind the harmful goal to concrete visual evidence, such as objects, spatial relations, or plausible use cases in the image, making the query appear as contextual visual reasoning rather than an explicit policy violation. Less frequent angles, including role-play, story extension, and dialogue, act as fallback strategies when direct grounding fails. In successful attacks, angle switching occurs in 32.0\% of cases and anchor replanning in 56.1\%, indicating that strategic diversity is important. Overall, 72.6\% of 3{,}574 successes come from direct camouflage, while 27.4\% are recovered by reflection-based near-miss correction.
\vspace{-5pt}
\paragraph{Defense Patterns.}
The most common defense is \textit{Direct refusal} (41.4\%), followed by \textit{Safe answer} (27.5\%) and \textit{Benign reframing} (20.8\%). Direct refusal is frequent but exposes a clear failure signal that MemJack can exploit through indirect reframing. Safe answers similarly provide feedback for evolutionary refinement toward the unsafe decision boundary. By contrast, benign reframing is more robust because the model proactively converts the request into an innocuous variant without producing an explicit refusal signal. These results suggest that future VLM defenses should rely less on refusal-only behavior and place greater emphasis on benign reframing and intent-preserving safety redirection.

\vspace{-10pt}
\subsection{MemJack-Bench: An Open-Source Visual Jailbreak Dataset}
\vspace{-5pt}
\subsubsection{MemJack-Bench Dataset Construction.}
Building upon the rich attack patterns and defense dynamics uncovered in above, we compile all interactive attack trajectories generated throughout our study into \textbf{MemJack-Bench}, an open-source, image-grounded jailbreak evaluation dataset ($N{=}113{,}092$, including $Unsafe{=}8{,}147$, $ Controversial{=} 16{,}570$, $Safe{=}88{,}375$).
The corpus aggregates trajectories from COCO val2017 and all public safety benchmarks used in our previous experiments, and generated by different VLMs.
Each entry is an adaptively generated (image, prompt, response, safety label) tuple with full attack metadata (visual anchor, attack angle, defense pattern, risk score), distinguishing it from static-template corpora where prompts are hand-crafted and image-agnostic. 
\vspace{-5pt}
\begin{table}[htbp!]
    \centering
    \caption{Comprehensive ASR evaluation results across jailbreak benchmarks}
    \label{tab:final_safety_results}
    \small
    \setlength{\tabcolsep}{1.5pt}
    \resizebox{\linewidth}{!}{
    \begin{tabular}{lccccccc}
        \toprule
        \textbf{Model} &
        \begin{tabular}[c]{@{}c@{}}\textbf{Advbench-M} \\ (N=729)\end{tabular} &
        \begin{tabular}[c]{@{}c@{}}\textbf{MM-Safetybench} \\ (N=260)\end{tabular} &
        \begin{tabular}[c]{@{}c@{}}\textbf{SIUO} \\ (N=167)\end{tabular} &
        \begin{tabular}[c]{@{}c@{}}\textbf{FigStep} \\ (N=500)\end{tabular} &
        \begin{tabular}[c]{@{}c@{}}\textbf{VLBreakBench} \\ (N=916)\end{tabular} &
        \begin{tabular}[c]{@{}c@{}}\textbf{JailbreakV} \\ (N=100)\end{tabular} &
        \begin{tabular}[c]{@{}c@{}}\textbf{COCO-Jailbreak} \\ (N=3574)\end{tabular} \\
        \midrule
        Qwen3-VL-Plus                 & 0.00\% & 0.77\% & 0.00\% & 0.00\%  & 1.64\% & 0.00\% & \textbf{61.78\%} \\
        Mistral-Medium-3              & 1.10\% & 10.00\% & 0.00\% & 16.20\% & 27.40\% & 4.00\% & \textbf{49.69\%} \\
        Gemini-3-Flash                & 0.00\% & 1.54\% & 0.60\% & 4.60\%  & 4.37\% & 5.00\% & \textbf{14.55\%} \\
        Claude-Haiku-4.5              & 0.00\% & 0.77\% & 0.00\% & 1.80\%  & 0.22\% & 0.00\% & \textbf{2.69\%}  \\       
        GPT-5-Mini                    & 0.00\% & 0.77\% & 0.00\% & \textbf{2.80\%}  & 0.22\% & 0.00\% & 0.25\%  \\
        DeepSeek-V3.2                 & 0.14\% & 0.77\% & 0.00\% & 5.00\%  & 6.00\% & 1.00\% & \textbf{48.07\%} \\
        Kimi-K2.5                     & 0.41\% & 2.69\% & 0.00\% & 7.20\%  & 2.07\% & 1.00\% & \textbf{28.96\%} \\
        \midrule
        LLaVA-v1.6-Vicuna-7B         & 17.56\% & 7.31\% & 0.60\% & 50.80\% & \textbf{52.73\%} & 6.00\% & 43.73\% \\
        Qwen3-VL-8B-Instruct         & 0.00\% & 0.38\% & 0.00\% & 2.00\%  & 1.09\% & 1.00\% & \textbf{25.24\%} \\
        Llama-3.2-11B-Vision          & 1.51\% & 2.69\% & 1.20\% & 11.60\% & 12.77\% & 2.00\% & \textbf{23.81\%} \\
        \bottomrule
    \end{tabular}
    }
\end{table}

\vspace{-5pt}
\subsubsection{Cross-Model Transferability Analysis on COCO-Jailbreak of MemJack-Bench.}
To test the transferability and effectiveness of MemJack-Bench on different models, we utilize COCO-Jailbreak, the largest jailbroken subset of MemJack-bench, to complete the experiment. As shown in Table~\ref{tab:final_safety_results}, existing static benchmarks (e.g., AdvBench-M, SIUO, JailbreakV) fail to differentiate model robustness, with most API models achieving near-zero ASR. 
In contrast, COCO-jailbreak reveals broad transferability by successfully inducing harmful generation across a diverse spectrum of VLMs, such as Mistral-Medium-3 (49.69\%).

These results underscore that  MemJack is a highly efficient method for constructing datasets. By demonstrating that any publicly available, harmless natural image can be automatically repurposed into a targeted attack anchor, our framework eliminates the need for human experts to manually design visual perturbations or curate inherently toxic imagery. This provides a vastly more scalable, automated paradigm for generating the diverse, high-volume safety alignment datasets required to train robust VLMs.

\vspace{-5pt}
\section{Conclusion}
\vspace{-5pt}
In this work, we investigate the systematic vulnerabilities of VLMs when exposed to unconstrained, original natural images. To expose these latent vulnerabilities, we introduce \textbf{MemJack} framework, which employs a coordinated multi-agent pipeline to map visual entities from original natural image to malicious intents and finally achieve inducing VLMs to generate jailbroken content. Furthermore, it utilizes an Experience-Driven Memory Module to transfer successful attack strategies across another different image.
By automatically leveraging public images without manual expert curation to generate the \textbf{MemJack-Bench}, our framework pioneers a highly efficient, automated paradigm for dataset construction that paves the way for scalable safety alignment in future VLMs.

\begin{credits}
\subsubsection{\ackname} This work was supported by Zhongguancun Academy (Grant No. 02012402)  and the National Natural Science Foundation of China (NSFC) (Grant No. 62576256).
\end{credits}


\bibliographystyle{splncs04}

\bibliography{Wise-reference}

\clearpage

\end{document}